\documentclass[twocolumn]{article}

\usepackage{amsmath} 
\usepackage{balance}
\usepackage{graphicx}
\usepackage[T1]{fontenc}
\usepackage{txfonts}
\usepackage{mathptmx}
\usepackage[pdflang={en-US},pdftex]{hyperref}
\usepackage{color}
\usepackage{booktabs}
\usepackage{textcomp}
\usepackage{todonotes}
\usepackage{url}
\usepackage{float}
\usepackage{subfigure}

\def\plaintitle{\textbf{Learning to Detect Touches on Cluttered Tables}}

\def\emptyauthor{}
\def\plainkeywords{
Interactive Tabletops; Touch detection; Hand interaction; Occlusion; Clutter}

\makeatletter
\def\url@leostyle{%
  \@ifundefined{selectfont}{
    \def\UrlFont{\sf}
  }{
    \def\UrlFont{\small\bf\ttfamily}
  }}
\makeatother
\def\pprw{8.5in}
\def\pprh{11in}

\definecolor{WildStrawberry}{rgb}{1.0, 0.26, 0.64}

\newcommand{\comment}[1]{}

\definecolor{linkColor}{RGB}{6,125,233}
\hypersetup{%
  pdftitle={\plaintitle},
  pdfauthor={\emptyauthor},
  pdfkeywords={\plainkeywords},
  pdfdisplaydoctitle=true, 
  bookmarksnumbered,
  pdfstartview={FitH},
  colorlinks,
  citecolor=black,
  filecolor=black,
  linkcolor=black,
  urlcolor=linkColor,
  breaklinks=true,
  hypertexnames=false
}

\begin{document}

\title{\plaintitle}


\author{
Norberto Adrian Goussies
\hspace{2mm}
Kenji Hata
\hspace{2mm}
Shruthi Prabhakara
\hspace{2mm}
Abhishek Amit$^*$
\\
Tony Aube$^*$
\hspace{2mm}
Carl Cepress$^*$
\hspace{2mm}
Diana Chang$^*$
\hspace{2mm}
Li-Te Cheng$^*$
\\
Horia Stefan Ciurdar$^*$
\hspace{2mm}
Mike Cleron$^*$
\hspace{2mm}
Chelsey Fleming$^*$
\hspace{2mm}
Ashwin Ganti$^*$
\\
Divyansh Garg$^*$
\hspace{2mm}
Niloofar Gheissari$^*$
\hspace{2mm}
Petra Luna Grutzik$^*$
\hspace{2mm}
David Hendon$^*$
\\
Daniel Iglesia$^*$
\hspace{2mm}
Jin Kim$^*$
\hspace{2mm}
Stuart Kyle$^*$
\hspace{2mm}
Chris LaRosa$^*$
\\
Roman Lewkow$^*$
\hspace{2mm}
Peter F McDermott$^*$
\hspace{2mm}
Chris Melancon$^*$
\hspace{2mm}
 Paru Nackeeran$^*$
\\
Neal Norwitz$^*$
\hspace{2mm}
Ali Rahimi$^*$
\hspace{2mm}
Brett Rampata$^*$
\hspace{2mm}
Carlos Sobrinho$^*$
\\
George Sung$^*$
\hspace{2mm}
Natalie Zauhar$^*$
\hspace{2mm}
Palash Nandy
\and
{Google Research, USA}
\\ $^*$Alphabetical Order
}
\date{}
\maketitle
\begin{abstract}
We present a novel self-contained camera-projector tabletop system with a
lamp form-factor that brings digital intelligence to our tables.
We propose a real-time, on-device, learning-based touch detection algorithm that makes
any tabletop interactive.
The top-down configuration and learning-based algorithm makes our method robust
to the presence of clutter, a main limitation of existing camera-projector
tabletop systems.
Our research prototype enables a set of experiences that combine hand
interactions and objects present on the table. A video can be found at \url{https://youtu.be/hElC_c25Fg8}.
\end{abstract}

\section{Introduction}

\begin{figure}[t!]
  \centering
  \subfigure[]{\includegraphics[width=.23\textwidth]{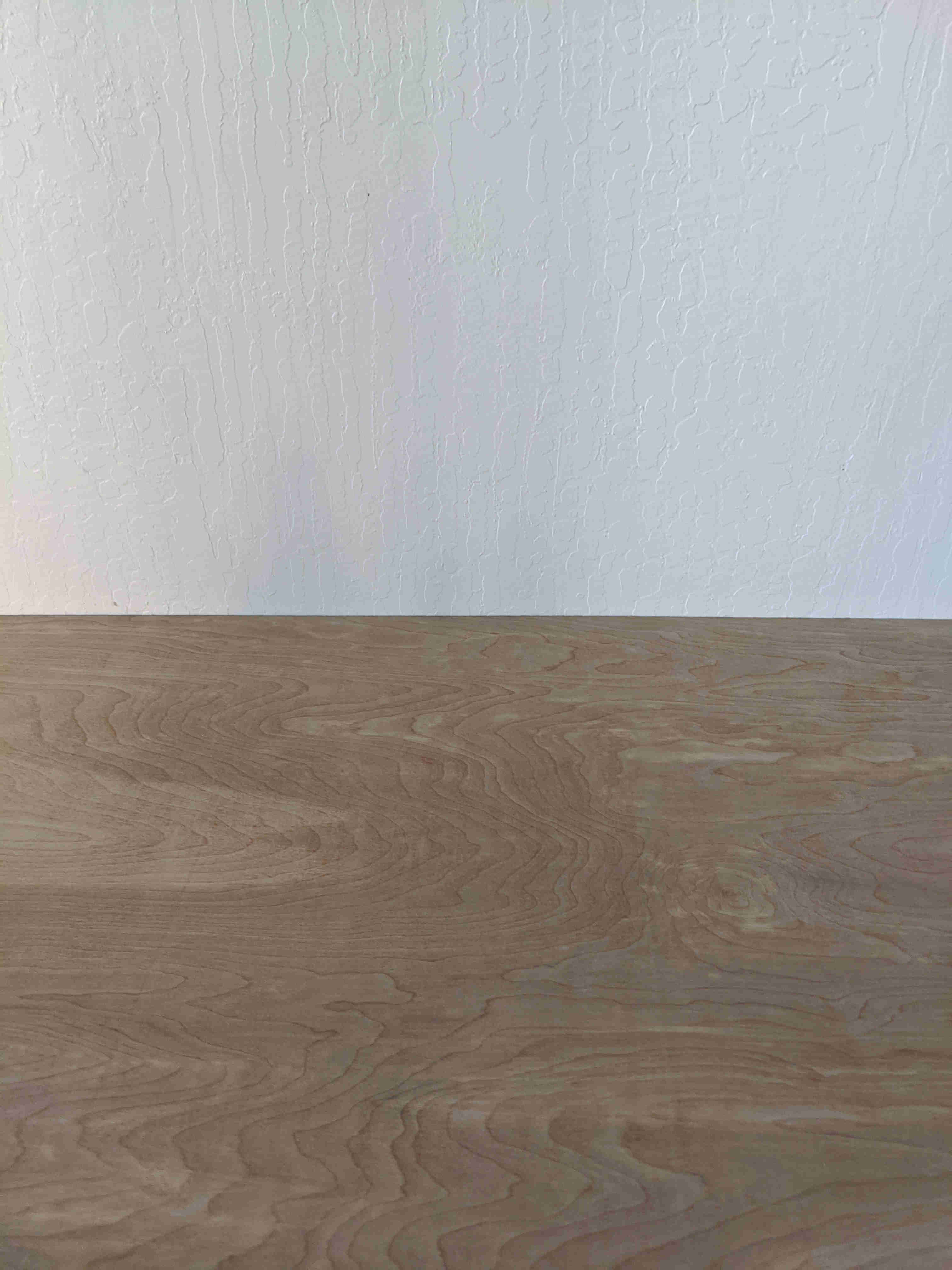}}
  \subfigure[]{\includegraphics[width=.23\textwidth]{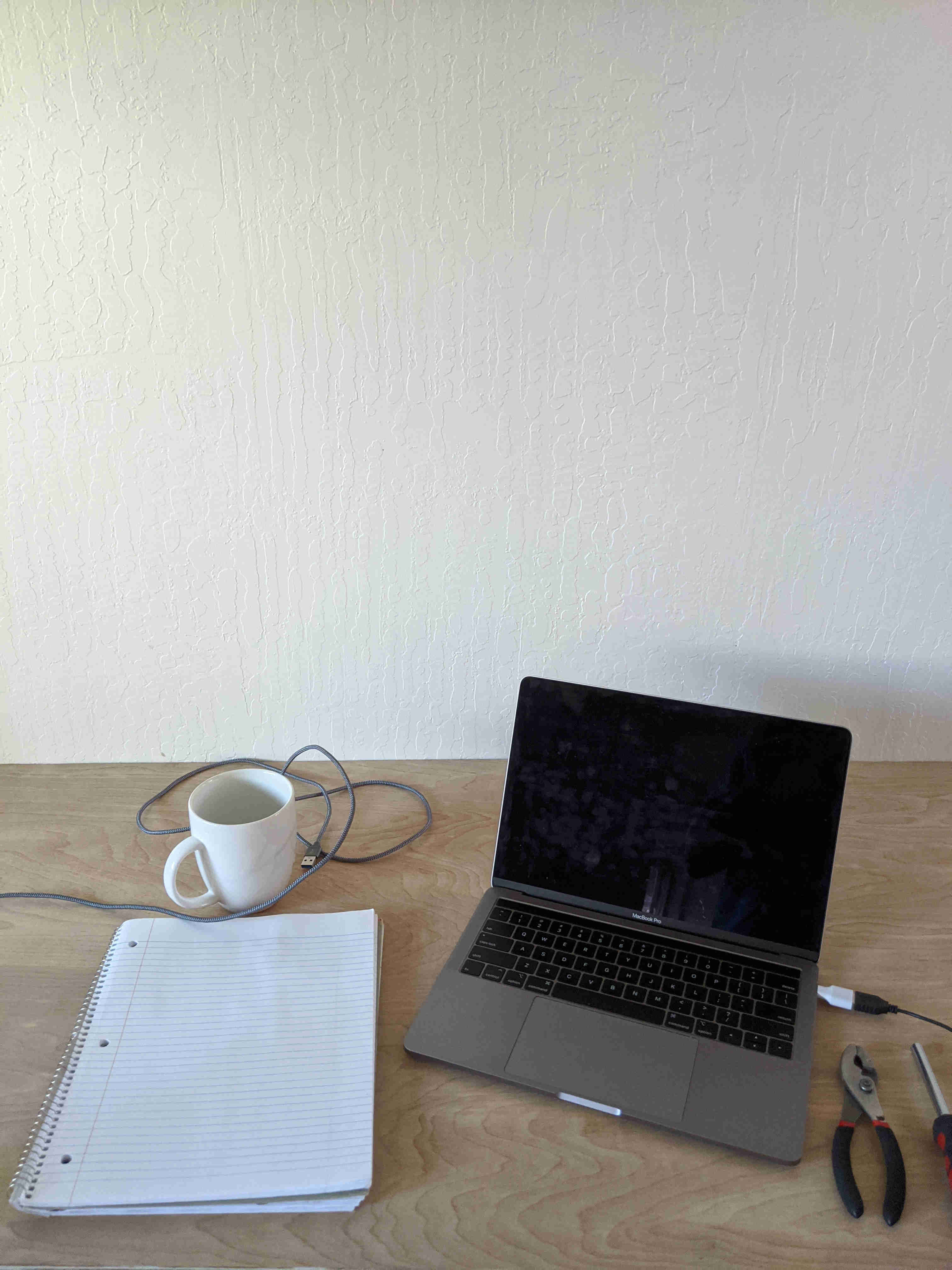}}\par
  \subfigure[]{\includegraphics[width=.23\textwidth]{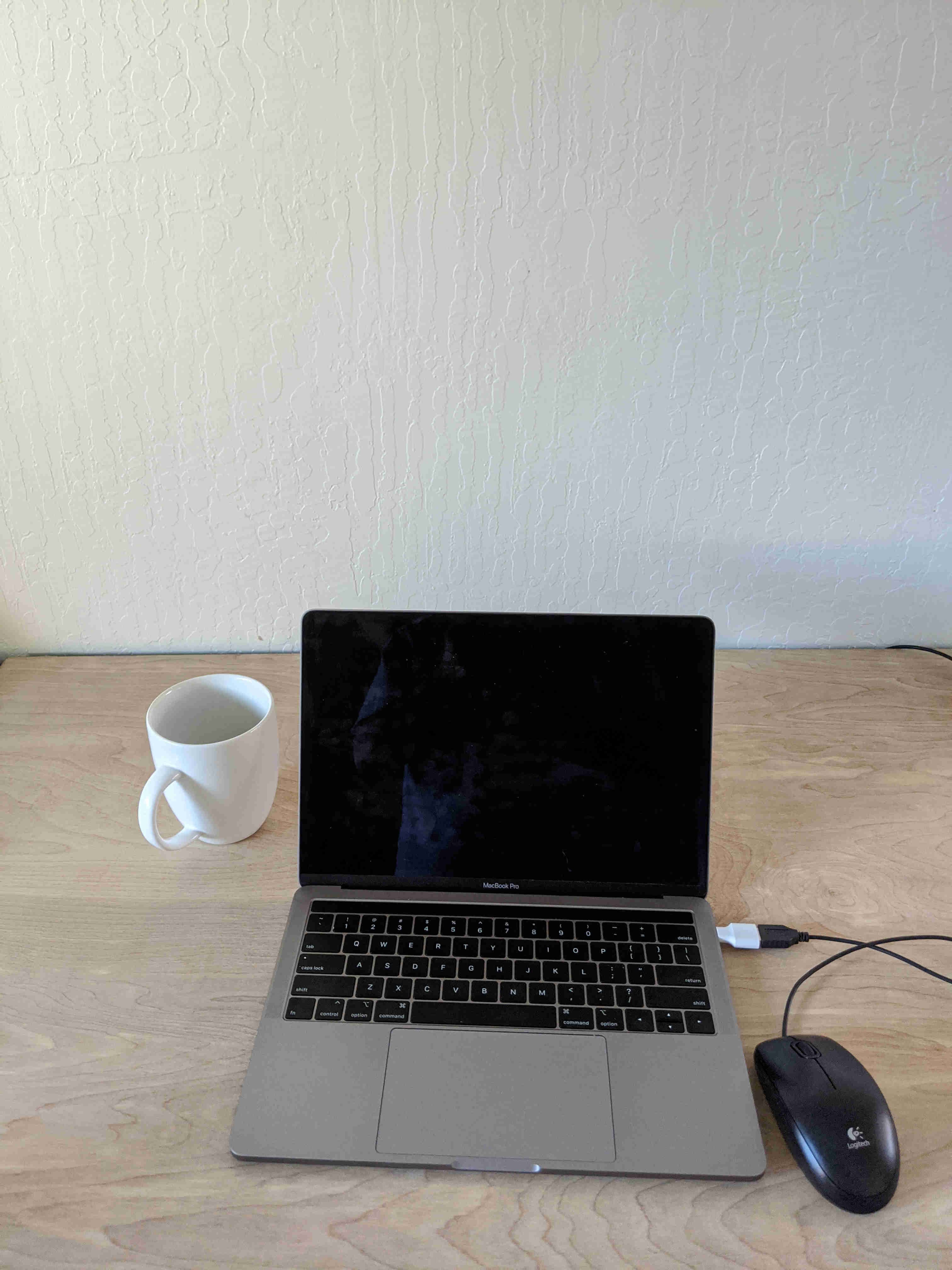}}
  \subfigure[]{\includegraphics[width=.23\textwidth]{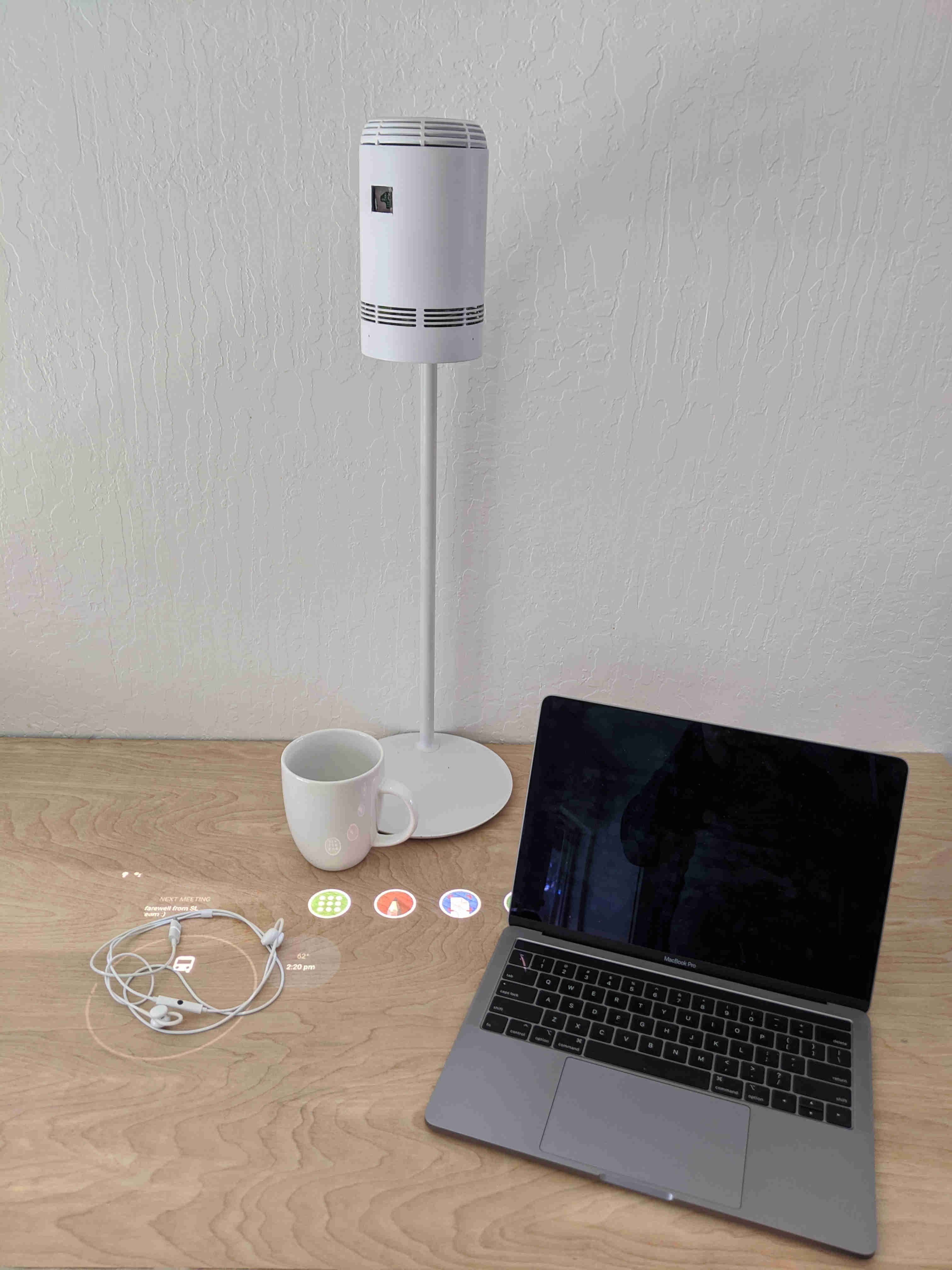}}
  \caption{Clutter comparison.
  (a) Tabletops in previous papers.
  (b) Our tabletop.
  (c) Typical tabletop.
  (d) Tabletops discussed in this paper.}
\label{fig:motivation}
\end{figure}

Interactive tabletop projectors are gaining attention, with many commercial and research prototypes being developed~\cite{2005playanywhere, 2016direct, 2018enhanced,  2018xperiatouch}. To provide a natural input mechanism, many of these tabletop projectors emulate a touchscreen by recognizing when human fingers contact the table surface.  The most common approach commercial devices follow is to project a sheet of
 infrared light just above the surface while watching for fingertips to intercept
 the light from just above. For example, the Canesta device \cite{2003full} is one of the first commercial devices to implement this approach.



To extend the interactions supported by these systems, several computer vision based
approaches have been proposed~\cite{2005playanywhere, 2010using,
2012extended, 2016direct}. These touch systems require the table surface to be mostly empty, which is not a common case. Therefore, the adoption of interactive tabletop projectors has been limited due to their inability to address challenges posed by the presence of clutter on a tabletop. Figure \ref{fig:motivation} illustrates a typical tabletop set up. In addition, 
these systems generally require expensive sensors or desktop machines with high-end GPUs and CPUs. Thus, these approaches are impractical for commercial devices that need to be robust to clutter on the table, portable, self-contained, low-cost, and have limited computational resources. 



\begin{figure}[h!]
\centering
\includegraphics[width=.45\textwidth]{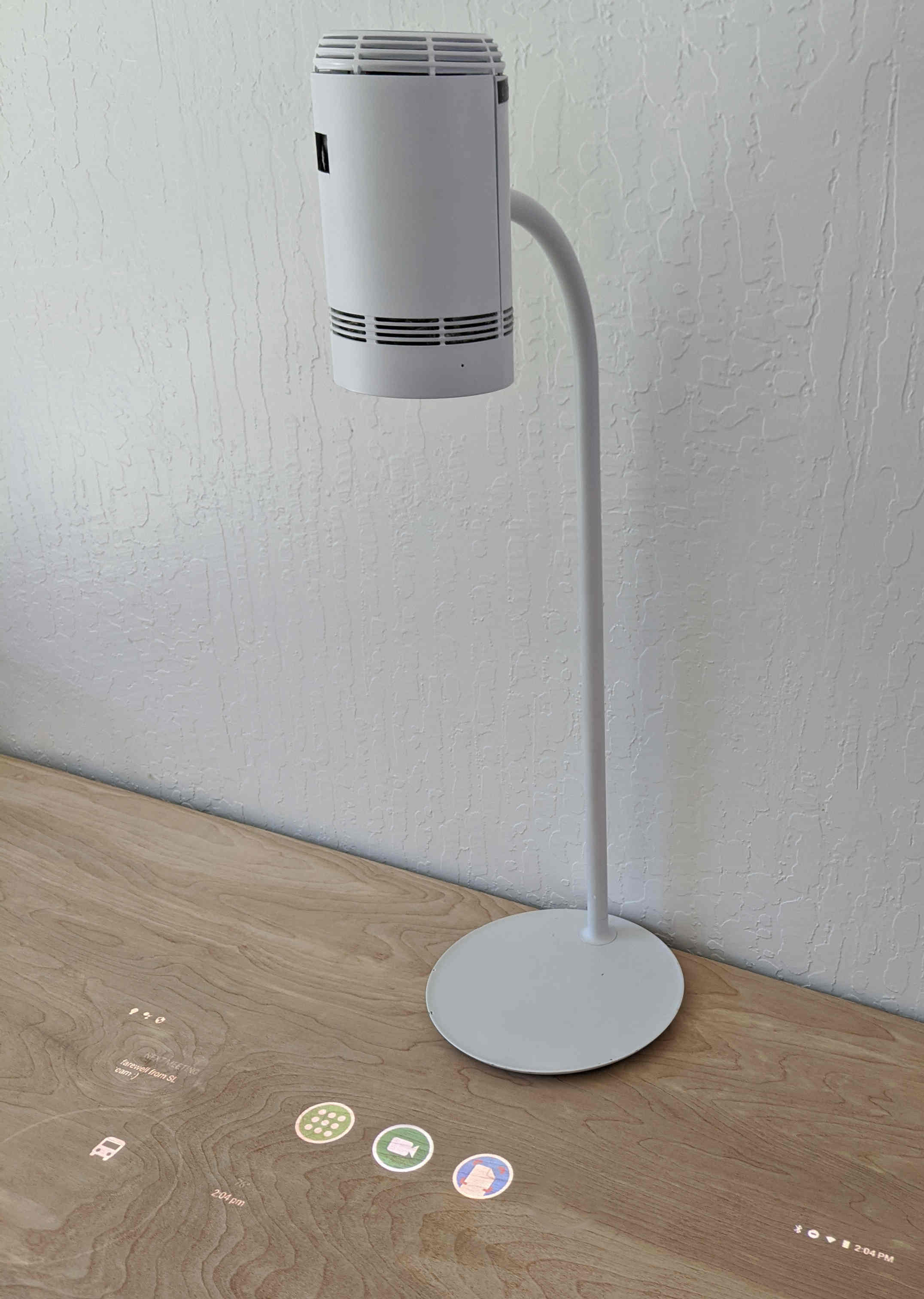}
\caption{Lamp form factor.}
\label{fig:lamp}
\end{figure}

In this paper we present a novel, low-cost, portable, and self-contained tabletop
projector prototype that has a lamp form factor (see Figure \ref{fig:lamp}).
To interact with the system we propose a novel real-time vision-based algorithm
that is based on encoding-decoding networks with skip connections
\cite{2016learning, 2017feature, 2016stacked}. Our algorithm uses stereo infrared
cameras to detect touches and in-air hand gestures.
The overhead cameras in the lamp make our prototype less prone to
occlusions, thereby overcoming an important limitation in most commercial
interactive tabletop projectors.
By being robust to clutter and occlusion, our system enables us to
build interactive tabletop projector experiences on realistic scenarios.
To our knowledge, this is the first time an end-to-end learning-based approach
has been proposed to detect when a finger is touching the table.

\section{Related Work}
\label{sec:related}




There is a large body of work on touch detection. The problem has received
attention since it has been a key component for interactive-tabletops
\cite{2016direct, 2010using, 2005playanywhere}, semi-transparent screens
\cite{2004touchlight}, and wearable devices \cite{2018mrtouch, 2011omnitouch}.

A wide range of cues have been explored to detect touches on un-instrumented
surfaces. Given the difficulty of the problem, initial approaches
\cite{1991digitaldesk, 2004touchlight} proposed relying on microphones to detect
touches. Later, PlayAnywhere \cite{2005playanywhere} proposed an algorithm based
on the analysis of shadows produced by the IR illuminant. PlayAnywhere's
front-facing configuration helps to overcome many occlusion problems, however
shadows are not easy to detect on a cluttered table.

Many approaches have been proposed to exploit the geometrical properties that
relate stereo cameras. Wren et al \cite{2001volumetric} proposed a technique
based on precomputed disparity maps, which can detect when a fingertip is
between an upper and lower plane without computing the entire depth range.
TouchLight \cite{2004touchlight} uses the homography induced by a plane
to detect when objects are near a semi-transparent screen. Visual Touchpad
\cite{2004visual} detects the fingertips in each camera independently and
applies a disparity threshold to decide if the finger is touching the touch-pad.
Finally, Agarwal et al \cite{2007high} proposed an algorithm that aggregates
stereo cues from several points at each fingertip by computing the intersection of the finger plane and the screen plane.

Compared with our approach, \cite{2001volumetric, 2004visual} use very low-resolution images and \cite{2007high} achieves only 20 fps.
Most of the approaches mentioned above are generally divided in two stages, first
they detect the fingertip and then they use a heuristic based on the stereo
images to decide if the fingertip is touching the surface.
Our end-to-end learning-based approach finds the fingertip position and decides
if the fingertip is touching the table in a single step. Our novel method runs
at 30 fps, is very accurate, and leverages the geometrical properties of stereo
cameras.

A recent trend is to leverage depth sensors to detect touches.
Wilson et al. \cite{2010using} explores a method to detect touches in which a
static background model is computed during initialization. Posterior depth
images are processed to identify areas with pixels that have a depth value
similar to the static background model. Murugappan et al. \cite{2012extended}
proposes to further refine the conditions to detect touches in \cite{2010using}
by using connected component analysis to discard blobs based on their area.
Additionally, a novel method for touch-gesture detection is proposed.
Xiao et al. \cite{2016direct} combines depth and infrared streams to obtain
precise finger boundaries, then examines a small depth pixel neighborhood around
the fingertip to decide if it is touching the table. Their method has a fallback
mode that allows them to cope with limited clutter conditions.
Cadena et al. \cite{2016fingertip} detects the arms using k-means and, then
finds the fingertips and refines their position using IR edge detection.
OmniTouch \cite{2011omnitouch} finds fingers by searching for cylinders in the depth image and performing flood filling to detect if it is touching a surface. Desktopography \cite{2017supporting} adds additional constraints to \cite{2011omnitouch} to reject finger-like objects.
Chai et al. \cite{2018enhanced} follows \cite{2010using} but the
fingertips are found using a neural network.
Fujinawa et al. \cite{2019occlusion} proposes a system in which
interactions are driven by hand postures.
The closest approaches to our work use neural networks to detect fingertips
or analyze hand pose \cite{2019occlusion, 2018enhanced}. The main difference
to our approach is that they rely on heuristics based on depth signals to decide
if a fingertip is touching the table.

A common limitation of these approaches is that they are not very precise in
disambiguating hover from touch. The main issue is that there are assumptions
about finger thickness. This assumption can be easily violated depending on
the angle the finger touches the surface. Additionally, finger thicknesses vary
depending on the user. In contrast, our approach can learn how to handle these situations from the data. By using stereo pair images, our neural network can
pick additional cues other than just the ones provided by the depth information.
Finally, it is important to note that even when depth
sensors have become widely available and they are not very expensive, they are
significantly more expensive than the stereo camera pair we use.


%

\section{Configuration}
\label{sec:configuration}

Figure \ref{fig:lamp} shows our lamp prototype which has two parts, a smart headlamp attached to a normal lamp base. The prototype is designed to sit on a flat
surface such as a table or desk. The head is 48 cm above the surface and contains
a vision module, projector and the main system
board where all the computing happens.
Apart from the power cable connected to the base, there are no other
cables or hardware required to use the lamp.

The vision module has two 1 MP infrared cameras (OV9282), with a 5cm baseline,
one infrared flood illuminator, and one 13MP RGB camera.
Similar to other camera-projector systems
\cite{2005playanywhere} we remove the projected graphics from the infrared
images using an infrared band-pass filter mounted in front of the infrared
cameras. In addition, we use BELICE dot emitters to add texture to the scene to
facilitate finding correspondences between the infrared cameras.
The vision module is similar to the one in the RealSense 435, but has a lower
cost since we do not need a dedicated ASIC to compute depth.

Our system has a projector based on the TI DLP4500/6401. The projector has a 1280x800
resolution and at 48cm over the surface has 71cm diagonal. The main
board has a Qualcomm 605 System-on-Chip (SoC), which supports up to three
cameras and can execute neural network inference on a built-in hardware accelerator
(Hexagon 685 DSP). The Qualcomm 605 processing capabilities are similar to the
ones of a Google Pixel 2.
The prototype runs Android Oreo and all the hand-user
interaction events are recognized and dispatched to the OS as if they were
coming from an input device. The total cost of the device is below the price of
most smartphones.

\section{Stereo Hand interaction}
\label{sec:handinteraction}

The stereo hand interaction module processes the frames from the two infrared
cameras to detect touches and in-air hand gestures. We use a two step process to
detect touches, first we detect the hands on the images from left camera, then
for each detected hand we use the stereo image pair to determine both the
fingertip locations and the subset that are touching the tabletop.

\subsection{Calibration and Table model}
\label{sec:tablemodel}

We perform a one-time calibration that estimates the intrinsic and extrinsic
parameters for all the cameras in the device, using a chessboard target and the
method described in \cite{2000flexible}. Then, we find the mapping between the
projector and the RGB camera by projecting a sequence of gray code patterns onto
the table \cite{2012simple}.

Our touch detection algorithm requires highly accurate estimation of the
distance between the infrared cameras and the surface the user interacts with.
A common solution when using depth sensors is to clear the surface and
build a static histogram using the first frames of the
depth sensor \cite{2018enhanced, 2010using}. Since requesting the user to empty
the table can be cumbersome, some methods maintain a rolling window of
the depth data \cite{2019occlusion, 2016direct}.

\begin{figure}[h]
\centering
\includegraphics[width=.45\textwidth]{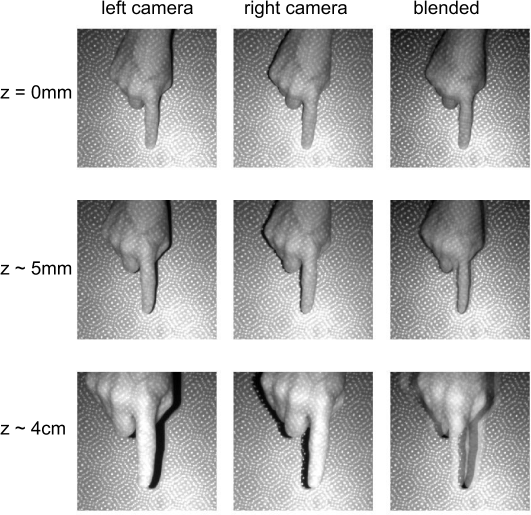}
\caption{Parallax induced by a plane.
First row: fingertip touching the table.
Second row: fingertip hovering a few millimeters above the table.
Last row: fingertip hovering a few centimeters above the table.
First column: images from the left camera.
Middle column: right images remapped using $D_x, D_y$.
Last column: left and remapped images blended together.}
\label{fig:parallax}
\end{figure}

However, we do not compute a depth map and instead infer the surface plane even through clutter, following a similar process to \cite{2004touchlight, 2007high, 2010touching}.
We capture the chessboard calibration pattern when it is lying flat
on the table and use all chessboard calibration corners to estimate the homography relating the
undistorted infrared images. Finally, we compute disparity maps
$D_{x}, D_{y} \in \mathbb{R}^{W \times H}$ that maps points on the
plane of the table from the left camera to the right camera.
Figure \ref{fig:parallax}
illustrates the image before and after the homography is applied.

\subsection{Hands Detection}
\label{sec:stageone}


The first step in our hand interaction algorithm is to detect hands in re-scaled
frames from the left camera
$I^{\textrm{small}}_{\textrm{left}} \in \mathbb{R}^{\frac{W}{F} \times \frac{H}{F}}$,
where $W$ is the width of the image, $H$ is the height, and $F$ is the re-scale
factor. Since the purpose of this step is to be able to obtain a rough estimate
of the hands position in the scene we heavily reduce the resolution of the left
image in this step using $F=8$.

Following \cite{2019objects}, our neural network predicts a
bounding box for each hand. Let $b = (x_1, y_1, x_2, y_2)$ be the coordinates of the
hand bounding box corners. The center of the bounding box is given by
$p = \left(\frac{x_1 + x_2}{2}, \frac{y_1 + y_2}{2}\right)$, and the size of the
bounding box is $s = (x_2-x_1, y_2-y_1)$.

Our neural network estimates
$\hat{Y} \in [0,1]^{\frac{W}{F \times R} \times \frac{H}{F \times R}}$,
where $R$ is the output stride. The heatmap $\hat{Y}$ predicts the center of the
bounding box. When $\hat{Y}_{\tilde{p_x}, \tilde{p_y}} = 1$ the neural network
is predicting a bounding box with center
$(\tilde{p_x}, \tilde{p_y}) = \left(\lfloor \frac{p_x}{R \times F} \rfloor, \lfloor\frac{p_y}{R \times F} \rfloor \right)$
and when $\hat{Y}_{\tilde{p_x}, \tilde{p_y}} = 0$ it is predicting that there is
no bounding box.
During training, for each low resolution bounding box center
$(\tilde{p}_x, \tilde{p}_y)$, we create the corresponding heatmap
$Y \in [0,1]^{\frac{W}{F \times R} \times \frac{H}{F \times R}}$ using a
Gaussian kernel $Y_{x,y} = \frac{1}{Z} \exp \left(- \frac{ (x-\tilde{p}_x)^2 + (y-\tilde{p}_y)^2 }{2 \sigma^2} \right)$,
where $\sigma^2$ depends on the size of the bounding box \cite{2014real} and
$Z$ is a constant to normalize the heatmap.
We define the loss for the center of the bounding box as:
\begin{equation}
L_{\textrm{center}} = \frac{-1}{N} \sum_{x, y}
\left\{ \begin{array}{lr}
        \left(1-\hat{Y}_{xy})^{\alpha} \log(\hat{Y}_{xy} \right) & \textrm{if } Y_{xy} = 1\\
        \left(1-Y_{xy})^{\beta} \hat{Y}_{xy}^{\alpha} \log(1- \hat{Y}_{xy} \right) & \textrm{otherwise}
        \end{array}\right.
\end{equation}
where $\alpha$ and $\beta$ are hyper-parameters of the focal-loss
\cite{2017focal} and $N$ is the number of hands in the image.

In addition, our neural network estimates
$\hat{S} \in \mathbb{R}^{\frac{W}{F \times R} \times \frac{H}{F \times R} \times 2}$.
We use $\hat{S}$ to regress the size of the bounding box, the value at the
center of the $\hat{S}$ is the predicted size of the bounding box.
We define the loss for the size as:
\begin{equation}
L_{\textrm{size}} = \left| \hat{S}_{p_x, p_y} - s \right|.
\end{equation}

The overall training objective is for the network is:
\begin{equation}
L_{\textrm{hand}} = L_{\textrm{center}} + \lambda_{\textrm{size}} L_{\textrm{size}}
\end{equation}
where $\lambda_{\textrm{size}}$ is used to change the importance $L_{\textrm{size}}$
in the overall training objective. In our experiments we fix
$\lambda_{\textrm{size}} = 0.1$.

\begin{figure}
\centering
\includegraphics[width=.45\textwidth]{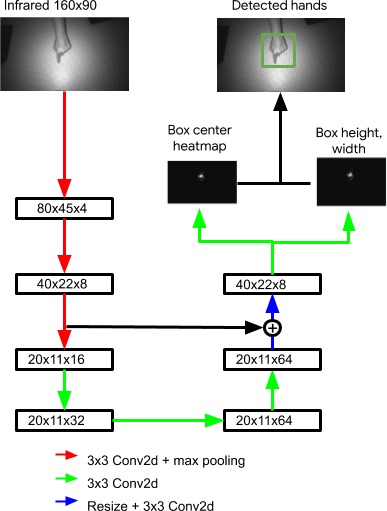}
\caption{Hand detection CNN architecture.}
\label{fig:firststage}
\end{figure}

The neural network architecture is shown in Figure \ref{fig:firststage}.
This is a fully convolutional neural network in which all layers are followed by
a batch normalization operation \cite{2015batch} and a ReLU non-linearity
(except for the last one) and has an encoder-decoder architecture with skip
connections\cite{2016learning, 2017feature, 2016stacked}.
Fast inference is central to our design and we have found that our
model is robust to many design choices. We experimented with more sophisticated
blocks \cite{2016deep} and observed minimal improvements.

The encoding layers of the network interleaves convolutional layers (with
point-wise non-linearities) and max-pooling layers to compute semantically
meaningful features. The pooling layers reduce the spatial resolution of the
features and introduce invariance to small spatial deformations.
The decoding layers of the network increase the feature resolution, using
bi-linear up-sampling and convolution layers \cite{2017devil}.
The resulting features are combined with features from the encoding layers via
skip connections. We decided to use these operations since they can be hardware
accelerated.
The output heatmaps $\hat{Y}$ and $\hat{S}$ share the same encoder-decoder
network. To produce each output heatmaps, the features of the encoder-decoder
network are passed through a separate $3 \times 3$ convolution.

We use random flips, rotations, brightness changes, and  contrast adjustment to
augment our training set. We use Adam \cite{2014adam} and set the learning rate
to $5e-4$ to optimize the overall objective from scratch.

\subsection{Touch Detection}
\label{sec:stereotouch}

Deep neural network techniques to estimate hand pose \cite{2014real} and person
pose \cite{2016stacked} have gained popularity due to the impressive performance
they have shown. RGB \cite{2017learning, 2018weakly} and depth \cite{2014real}
are the most common input modalities.
The number of joints to represent hand pose vary depending on the datasets
that are used for evaluation; \cite{2014real} uses 14 joints, \cite{2014latent}
uses 16 joints, and \cite{2014realtime} uses 21 joints.
The predicted joints are either represented using heatmaps or using their
2D coordinates.

In our system, instead of predicting the full hand pose we train a neural network
that only predicts the position of the fingertips
$k^{1}, \ldots, k^{5} \in \mathbb{R}^2$ and the palm $k^6 \in \mathbb{R}^2$,
which are the only keypoints that we need.
Our neural network estimates three heatmaps:
$\hat{K}^{\textrm{fingertips}} \in [0,1]^{\frac{W}{Q} \times \frac{H}{Q}}$
encodes the position of all the fingertips in the hand,
$\hat{K}^{\textrm{touch}} \in [0,1]^{\frac{W}{Q} \times \frac{H}{Q}}$
encodes which fingertips are touching the table, and
$\hat{K}^{\textrm{palm}} \in [0,1]^{\frac{W}{Q} \times \frac{H}{Q}}$ encodes the
position of the palm, where $Q$ is the output stride for the neural network.

During training, we remap the $I_{\textrm{right}} \in \mathbb{R}^{W \times H}$
using linear interpolation:
\begin{equation}
I_{\textrm{remapped}} = I_{\textrm{right}}(D_x(x,y), D_y(x,y)).
\end{equation}
We feed $I_{\textrm{remapped}}$ and $I_{\textrm{left}}$ to our neural network.
As in the previous section, we use Gaussian
kernels to create the heatmaps. For the fingertips heatmap
$K^{\textrm{fingertips}}$ we use only one heatmap to encode the position of all
the fingertips since we are not interested in which specific finger is touching:
\begin{equation}
K^{\textrm{fingertips}}_{x, y} =
\max_{ i \in {1, \ldots, 5}} \frac{1}{Z_i}
\exp \left(- \frac{ (x-k^i_x)^2 + (y-k^i_y)^2 }{2 \sigma^2} \right)
\end{equation}
where $Z_i$ is a normalizing constant and $\sigma$ is a fixed parameter for all
keypoints.
We create the touching heatmap $K^{\textrm{touch}}$ in the same way we create
the fingertips heatmap $K^{\textrm{fingertips}}$ but we only consider the
fingertips that are touching the table. If a keypoint is not visible we ignore
it when we create the heatmaps. We use the Mean Square Error (MSE) loss
\cite{2014real, 2016stacked} to compare
the predicted heatmaps with the ground-truth as training objective:
\begin{equation}
\begin{array}{ll}
L &= \frac{1}{W \times H}\sum_{x,y} \left(K^{\textrm{fingertips}}_{x,y} - \hat{K}^{\textrm{fingertips}}_{x,y}\right)^2 \\
  &+  \frac{1}{W \times H}\sum_{x,y} \left(K^{\textrm{palm}}_{x,y} - \hat{K}^{\textrm{palm}}_{x,y}\right)^2 \\
  &+ \frac{1}{W \times H}\sum_{x,y} \left(K^{\textrm{touch}}_{x,y} - \hat{K}^{\textrm{touch}}_{x,y}\right)^2.
\end{array}
\end{equation}

During inference, for each detected hand bounding box $b$, we crop the left image
$I^{b}_{\textrm{left}}$ and use the disparity maps $D_x, D_y$ to remap
the corresponding bounding box in the right image $I^{b}_{\textrm{remapped}}$.
We feed $I^{b}_{\textrm{left}}$ and $I^{b}_{\textrm{remapped}}$  images to our
neural network.
Then, we decode the fingertip positions $\hat{k}^1, \ldots, \hat{k}^5$
by finding the local maximas (and using non-maximal suppression) in
$\hat{K}^{\textrm{fingertips}}$. Once we have the local maxima coordinates we
fit a 2D Gaussian function to obtain the coordinates with sub-pixel precision.
To decide if a fingertip $i$ is touching the table, we take the maximum value
of $\hat{K}^{\textrm{touch}}$ in a window centered around the fingertip position
$\hat{k}^i$.

\begin{figure}
\centering
\includegraphics[width=.45\textwidth]{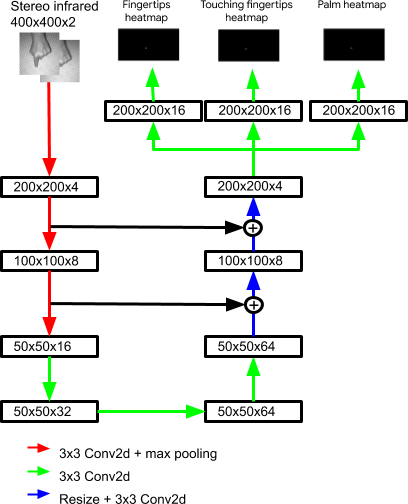}
\caption{Fingertip, palm, and touch detection CNN architecture.}
\label{fig:secondstage}
\end{figure}

Figure \ref{fig:secondstage} shows the neural network architecture for this
stage, which is similar to the neural network to detect hands but with more
decoding layers. The additional decoding layers enable us to obtain spatially
accurate positions for the predicted keypoints.
It is important to note that, since our neural network is fully convolutional,
the image sizes that we use to train the network do not need to be the same as
the ones that we use when we run inference.
We train using full stereo images but we feed cropped stereo images
during inference.

The main reason our neural network can estimate $\hat{K}^{\textrm{touch}}$
is that it can exploit geometrical constraints and other cues.
A fingertip touching
the table should have a similar position in $I_{\textrm{left}}$ and
$I_{\textrm{remapped}}$. The further away a fingertip is from the table,
the further apart its position is in $I_{\textrm{left}}$ and
$I_{\textrm{remapped}}$.
Figure \ref{fig:parallax} illustrates this property.
By using a learning-based approach that recognizes
when a finger is touching the table, we are able to successfully handle many
cases that were found to be difficult in previous heuristic-based approaches
\cite{2018mrtouch, 2018enhanced}.
For example, when a finger is touching the table perpendicularly,
heuristics that use connected components struggle to detect touches because the
connected component that the finger creates is small. However, our system easily
handles this case.

\subsection{Touch Dataset}

To train the touch detection CNN we need a dataset with accurate labels of when
the fingertips are touching the table. The main challenge is that annotating
whether a fingertip is touching the table using stereo images is time consuming
and error prone. Therefore, when we collected the
dataset we requested the participants to perform tasks in which we knew which
fingers were touching the table and which fingers were hovering over the table.

We collected 50 thousand stereo image pairs from 13 volunteers and 3 different
devices.
Each volunteer was requested to perform 14 tasks for 20 seconds each. The tasks
consisted of moving one hand around the projected area
while performing a gesture with a pre-determined subset of fingers from that
hand. For example, some of the gestures were:
touch while pinching with the thumb and index finger, touch while performing two
finger scroll,
and move the hand above the table without touching it. Also, we  requested the
participants to hover their hand above the table but without touching it, while
performing similar gestures, in order to obtain samples that can help us learn
an accurate touch detector.
We tried to emulate realistic tabletop scenarios by introducing a variety of
clutter into the scene.
Between different tasks, we modified the scene by adding
or removing objects, as well as changing the illumination or the surface material.
After collecting the stereo images we annotated the keypoint locations on the
left image.

\subsection{Multi-Touch Tracking}
\label{sec:tracking}

Our system can detect up to two hands and ten contact points simultaneously.
Given two consecutive frames at time $t$ and $t+1$, and the detected contact
points set $D_t$ and $D_{t+1}$ in each frame, we match the contact points
$p \in D_{t+1}$ to $q \in D_{t}$ if the following
two conditions hold simultaneously:
\begin{equation}
q = \underset{q' \in D_t}{\mathrm{argmin}} \|p - q'\|_2^2
\end{equation}
\begin{equation}
p = \underset{p' \in D_{t+1}}{\mathrm{argmin}} \|q - p'\|_2^2
\end{equation}

When this happens, we report a drag event if the $\|p - q\|_2^2 > 2.5$ mm.
When a contact point $p \in D_{t+1}$ does not match to any contact point in $D_{t}$,
we report the contact point $p$ as a touch down event to the Android OS.
Similarly, when a contact point $q \in D_{t}$ does not match to any contact
point in $D_{t+1}$ nor $D_{t+2}$ we report a touch up event.
By reporting events to the Android OS, and keeping a stable fingertip identity
among frames, many of the Android gestures
(one or two finger drag, long press, pinch-zoom, etc...) work well, allowing us
to use most Android applications without modifications.

\subsection{In-Air Hand Gesture Detection}
\label{sec:stereotouch}
\begin{figure}[h]
\centering
\includegraphics[width=.45\textwidth]{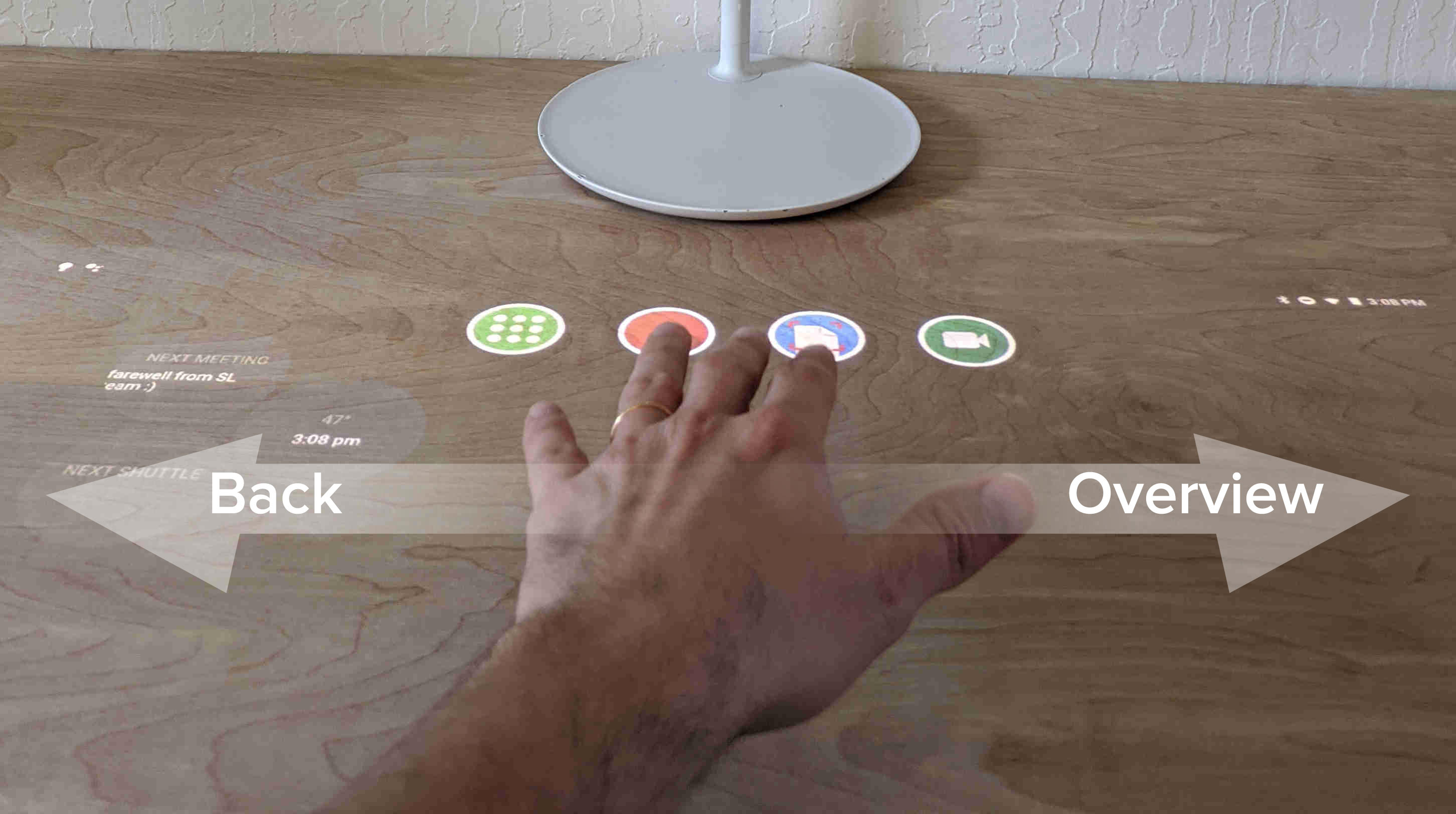}
\caption{Navigation gestures.}
\label{fig:gestures}
\end{figure}

We use in-air gestures to replace the Android navigation bar.
Our system detects two in-air gestures: back and forward, as depicted in
Figure \ref{fig:gestures}. The gestures are mapped to the back and overview buttons in the Android
navigation bar. In order to detect in-air gestures we implemented heuristics
based on speed and directionality of the user's hand motion.

\section{Implementation Details}
\label{sec:implementation}

The CNNs in the system are run on Qualcomm's DSP using TF-Lite's hexagon
delegate and are trained using quantization-aware training
\cite{2018quantization}. The neural network in the first stage takes 6ms to run
and the one in the second stage takes 12ms per hand, we process two hands at
most and ignore the rest of the hands in the scene. By offloading inference to Qualcomm's DSP, and using only one slow SoC core for the touch detection algorithm, we can efficiently distribute the compute workloads on the device. The big cores and most of the small cores on the SoC are reserved for running the Android operating system and applications.
\begin{figure}
\centering
\includegraphics[width=.45\textwidth]{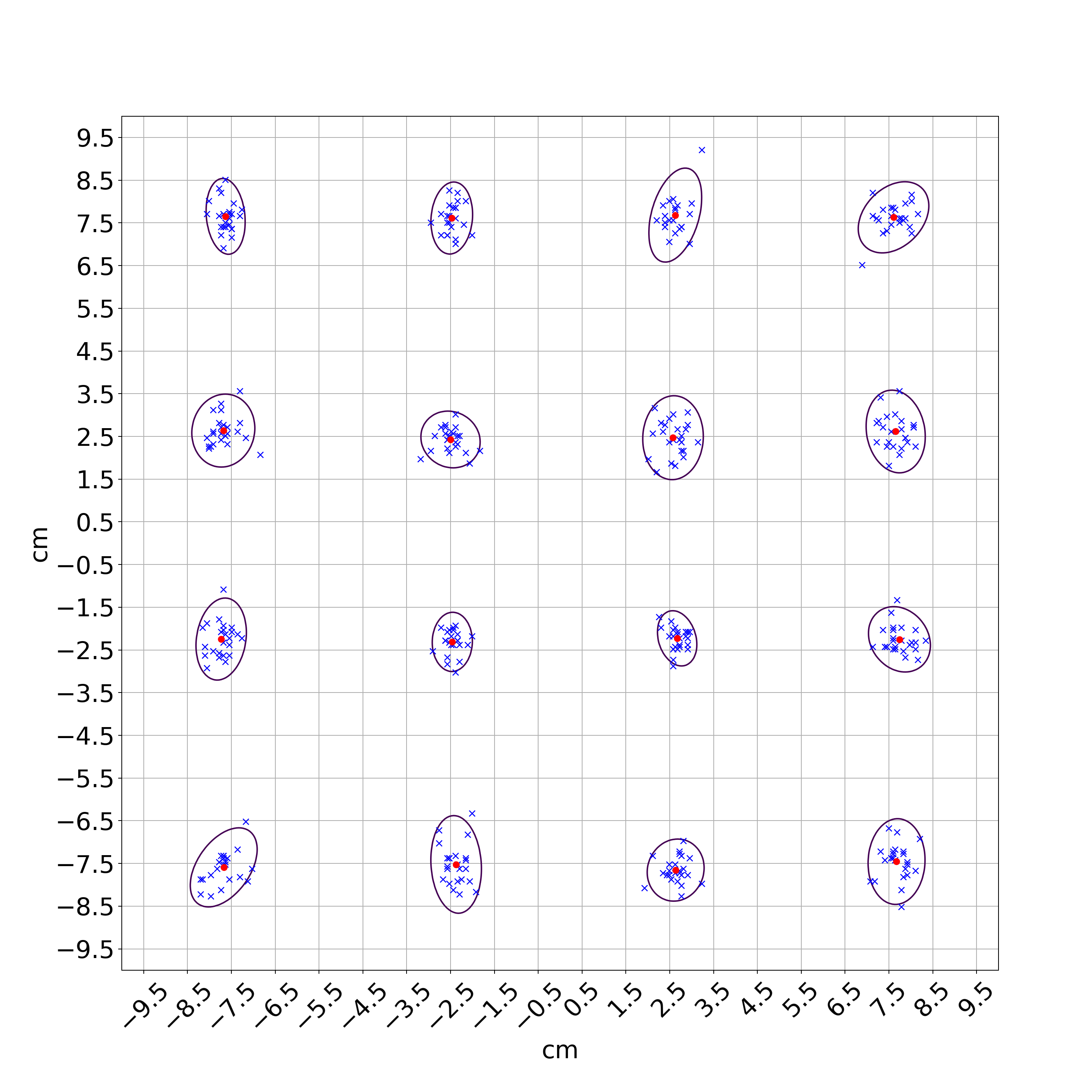}
\caption{Scatterplot of touch points, plotted with 95\% confidence intervals.}
\label{fig:touchscatter}
\end{figure}

\section{Experiments}
\label{sec:experiments}

We conducted experiments with our system through an evaluation of touch accuracy,
latency measurements, dealing with failure cases, and informal testing with
various experiences.

\subsection{Touch Accuracy}
To analyze the spatial and touch accuracy of the system we performed a user
study similar to the cross-hair targeting study described in \cite{2018mrtouch}.
We requested 12 participants to tap once on the
cross-hair targets projected on the table. A total of 16 cross-hair targets were
projected. All users were given a training period of 10 minutes.

\begin{figure}
\centering
\includegraphics[width=.45\textwidth]{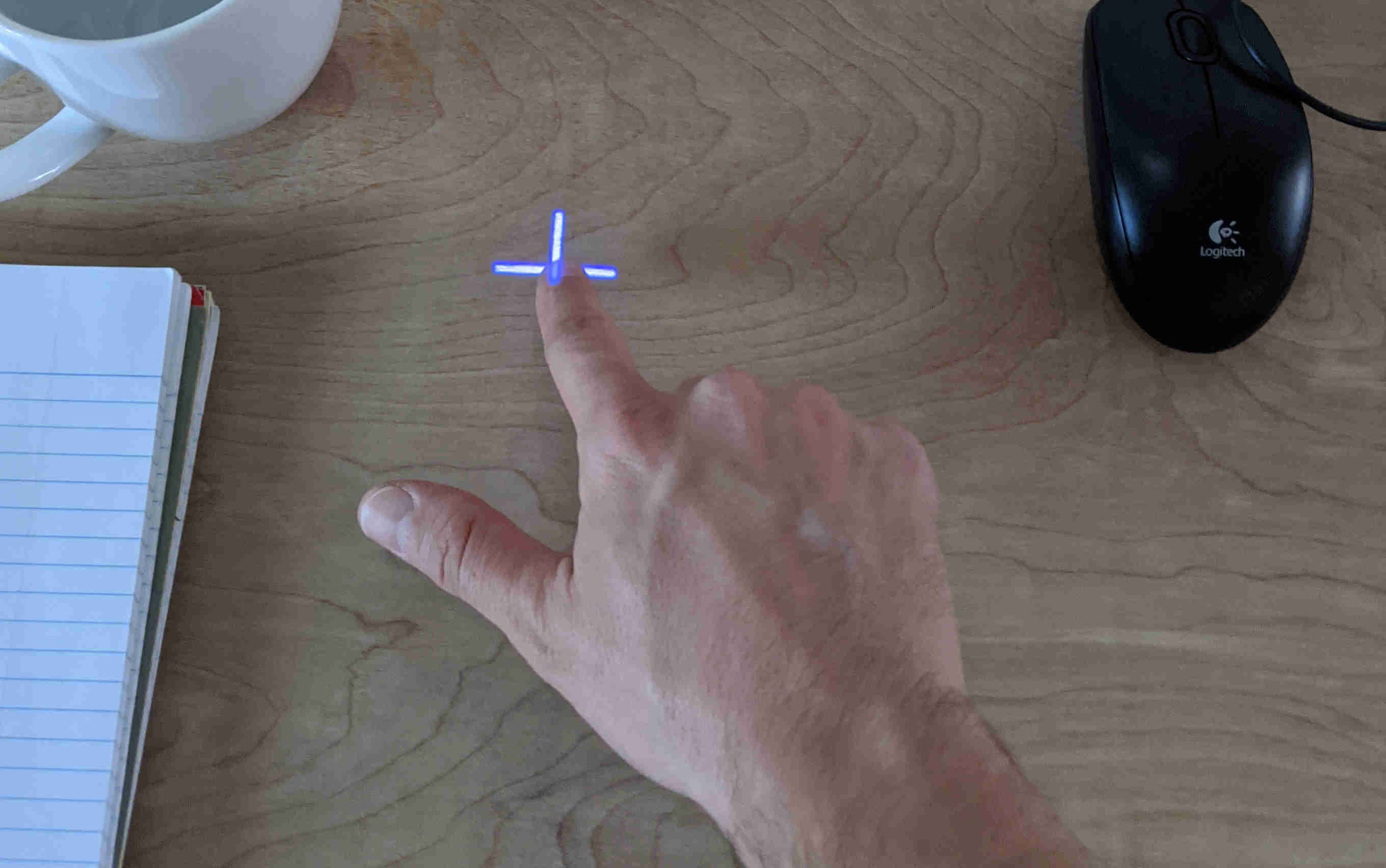}
\caption{Example of the cross-hair target task while clutter is present on the
table.}
\label{fig:task}
\end{figure}

The targets were placed in a $4 \times 4$ grid evenly spaced across a
$20 \times 20$ cm square area. The targets were displayed in random order, one
at the time, for three seconds each. The system recorded all the touch events.
To capture diversity of interactions, we evaluated under two different
scenarios, with the user seated in a chair, and with the user standing.
Clutter was present on the table, but not in places where the target was shown
(See Figure \ref{fig:task}).

Across the 12 participants, we obtained
384 cross-hair trials. Of these, 10 (2.6\%) reported
no touch contact, 370 (96.35\%) exactly one contact and 4 (1.04\%) more than
one contact. In the event of multiple detected contacts, we decided to ignore
the trial for the rest of the analysis. We found no significant
spatial or touch accuracy differences between the user in seated or standing
position.
Figure \ref{fig:touchscatter} plots the touch points and the 95\% confidence
intervals for the seated and standing positions. 98\% of touch interactions were detected within 1 cm of the actual
target. The average Euclidean error is 0.45cm.

\subsection{Comparison with Existing Approaches}

\comment{
\begin{figure}
\centering
\includegraphics[width=.45\textwidth]{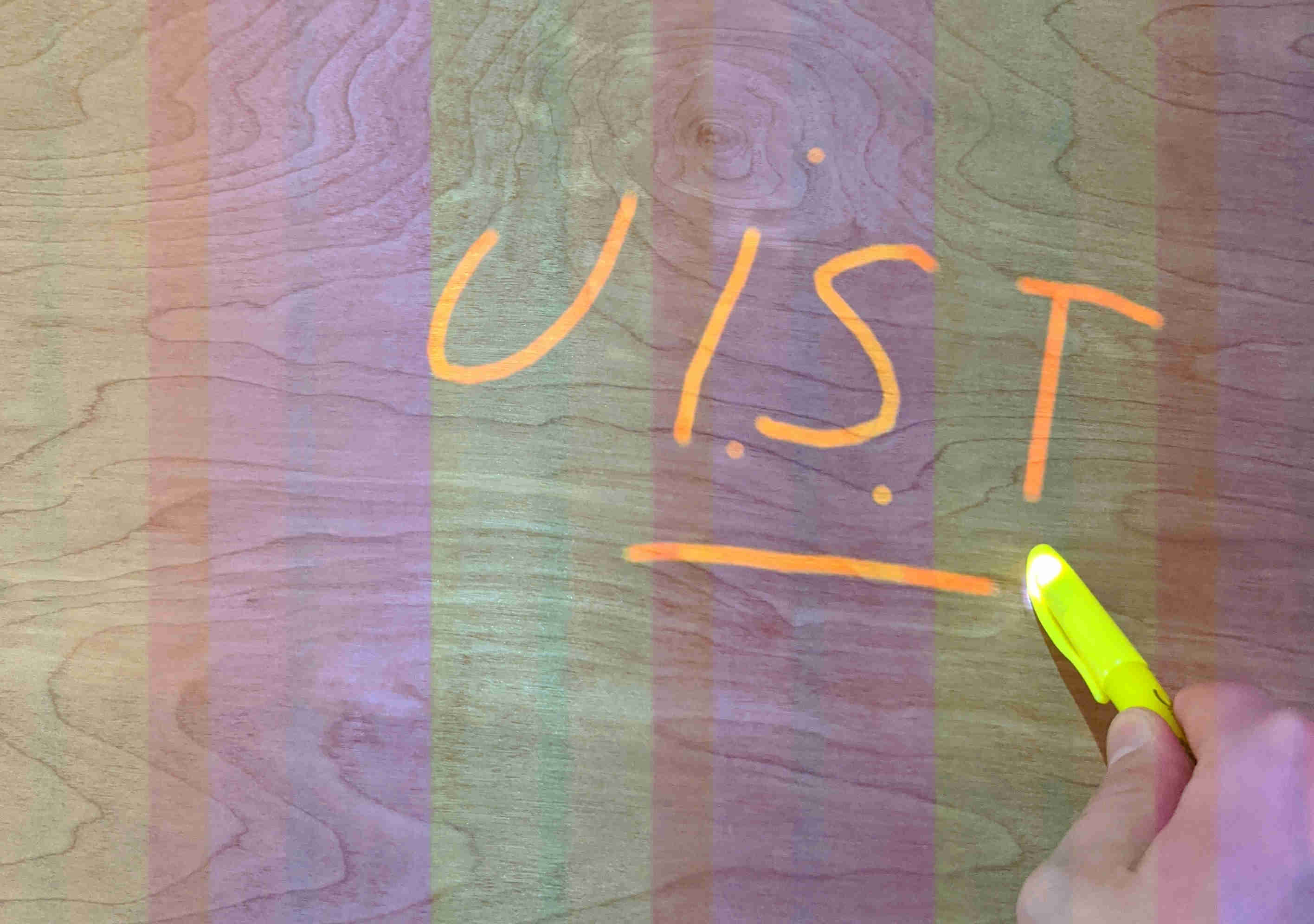}
\caption{Failure case example: Pens are detected as fingers.}
\label{fig:fail}
\end{figure}
}

\begin{table*}[]
    \centering
\begin{tabular}{c|c|c|c|c|c|c|c}
& Camera  & Sensor  & Avg. Euclidean & Touch Detection  & Clutter  & Hardware & Cost \\
& Distance  & Type  & Error & Rate &  Robustness &  &  \\
\hline
\hline
Wilson \cite{2010using} & 75cm & depth & 7mm & N/A & low & Desktop machine & medium \\
 &  &  structured-light & &  &  & CPU  &  \\
\hline
Xiao et al \cite{2016direct} & 160cm & IR+depth & 4.8mm & 99.3\% & medium & Desktop machine & medium \\
 &  &  time-of-flight & &  &  & CPU  &  \\
\hline
Chai et al \cite{2018enhanced} & N/A & depth & 2.5mm & N/A & low & Desktop machine & high \\
 &  &  structured-light & &  &  & high-end GPU  &  \\
\hline
Xiao et al \cite{2018mrtouch} & 51cm & IR+depth & 5mm & 72\% & medium & HoloLens & low \\
table condition &  &  time-of-flight & &  &  &  &  \\
\hline
Ours & 48cm & stereo IR   & 4.9mm & 96.35\% & high & Qualcomm  & low \\
 &  &  structured-light & &  &  & 605 SoC &  \\
\hline
\end{tabular}
\caption{Comparison of touch  approaches with ours.}
\label{tab:comparison}
\end{table*}

There are many touch detection algorithms that have been proposed in the past.
Table \ref{tab:comparison} shows a comparison between our approach and closest previous systems. We compare with approaches in which the setup is similar to
our top-view interactive tabletop projector system. 
There is a wide variety of setups, costs, camera distances, and results. 
Our approach achieves the best touch detection rate of the low cost systems and it is the most robust to clutter.

To obtain a rough cost estimate for each approach we considered that desktop machines are more expensive than embedded devices and that high-end GPUs are extremely expensive. Time-of-flight depth sensors are more expensive than structured-light depth sensors. Stereo infrared cameras with structured-light are the cheapest alternative.

To analyze the clutter robustness of each system we considered whether the method supports dynamic backgrounds and, if it does, how complex the scene can be. For example, \cite{2016direct, 2018mrtouch} have problems to detect touches in the presence of documents like the one in Figure \ref{fig:dmv} since it is very likely that an overfill error gets triggered.

\subsection{Latency}
\label{sec:latency}
We measured the end-to-end latency of the system, defined as the time between the
physical interaction and the first frame that incorporates it. This measurement
includes the time required to capture the frames, process them and dispatch the
OS event.

\begin{figure}
\centering
\includegraphics[width=.45\textwidth]{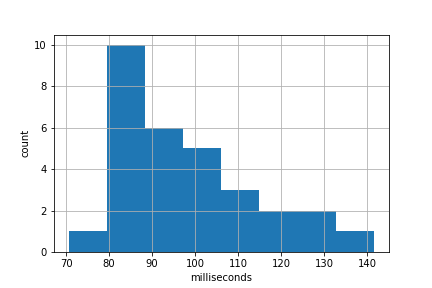}
\caption{Touch system latency distribution.}
\label{fig:latency}
\end{figure}

To measure the end-to-end latency of the system, we positioned a high-speed
camera near lamp base and we recorded 30 tap interactions with our system.
Figure \ref{fig:latency} shows the latency distribution of the system.
We observed a mean latency of our touch system of $98$ ms and a mode of $91$ ms.
Out of these 98 ms, 31 ms are spent in the touch system. The remaining 67 ms
are spent in the rest of pipeline and there are many opportunities for
optimizing it.

As mentioned  in \cite{2018mrtouch} latency is crucial for touch inputs and
is more easily noticed in large displays because users can move their hands faster.
Our latency compares favorably to the 180 ms latency reported by
\cite{2018mrtouch} and even some touchscreen latencies, which are in the
(50-200ms) range \cite{2012designing}.

\subsection{Failure Cases}

We evaluated our system on a comprehensive set of use-cases.
While we have significantly improved upon the existing use-cases, our system
still fails in a few edge cases.
One such example is objects resembling a finger can
occasionally trigger false positives. We could in
the future augment the models with RGB images (in addition to infrared) to help
disambiguate these cases.
Another failure case is miss-detection when user touches are at the edge of
camera's field of view.
The main reason is that the hand is only partially visible in the infrared
cameras.
This could be mitigated by using a camera with a field of view that is wider
than area of interest on a tabletop.

\begin{figure}
\centering
\includegraphics[width=.45\textwidth]{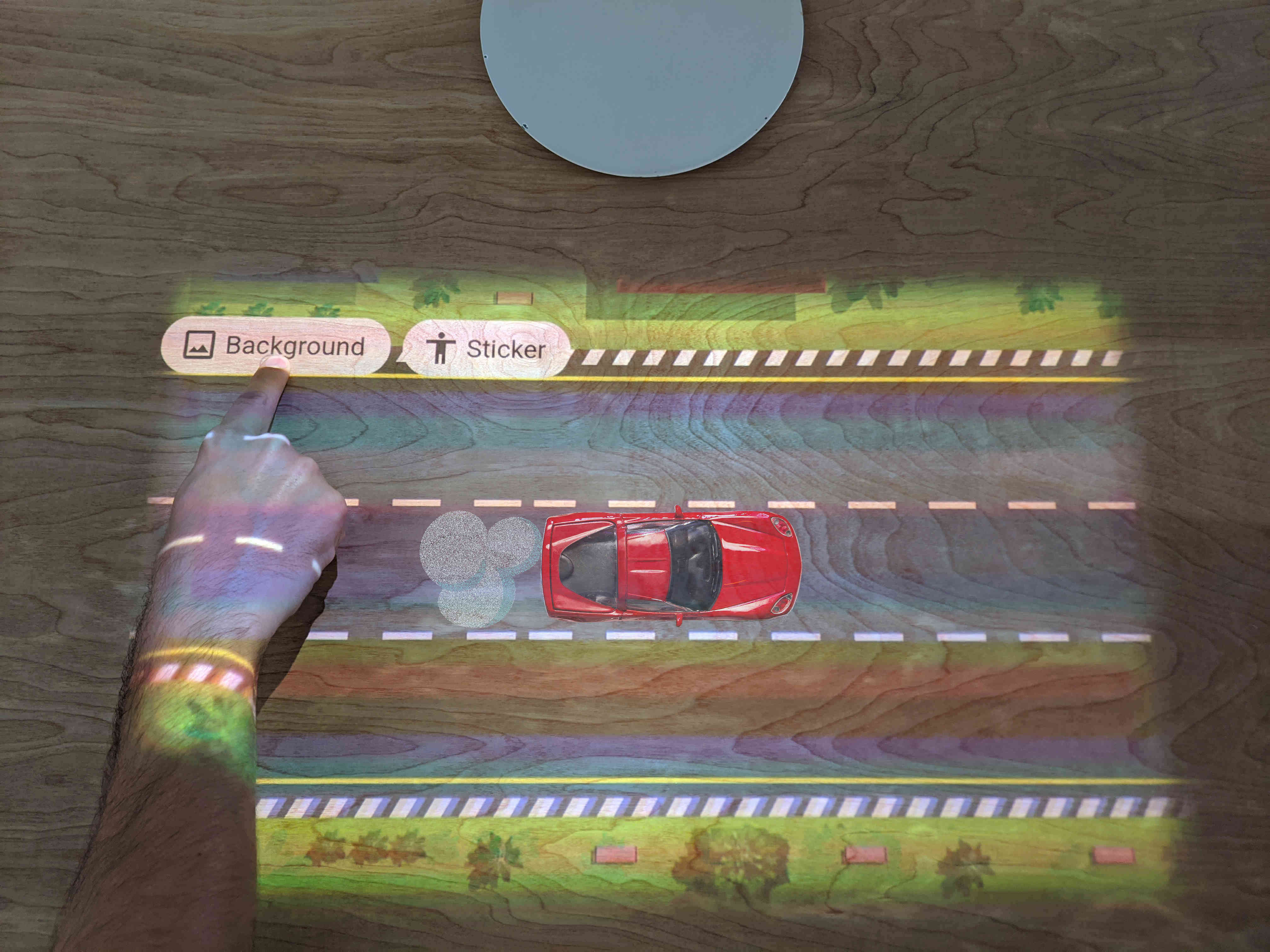}
\caption{Sandbox demo application.}
\label{fig:sandbox}
\end{figure}


\subsection{Sandbox}

We developed a demo application, Sandbox, that allows the user to choose from
different dynamic backgrounds that are projected on toys to augment them with
animations, see Figure \ref{fig:sandbox}. The toys can be moved through the
projected area and animation will track them. The example shows a moving animated race track background, giving the illusion that the toy car is moving. The background can react to the
position of the detected toy cars and present different obstacles, turning the table into a race game with real toys. In this example, we also project animated smoke behind the car, amplifying the illusion of movement.

In order to detect when a new toy is present, the Sandbox application detects
when the user's hand enters or leaves the projected area. By comparing the frames
associated to those events, we determine if a new object has been placed on
the table. We then use the frame the toy is detected in as a reference to track it
in following frames using optical flow.

\subsection{Desktop}

\begin{figure}
\centering
\includegraphics[width=.45\textwidth]{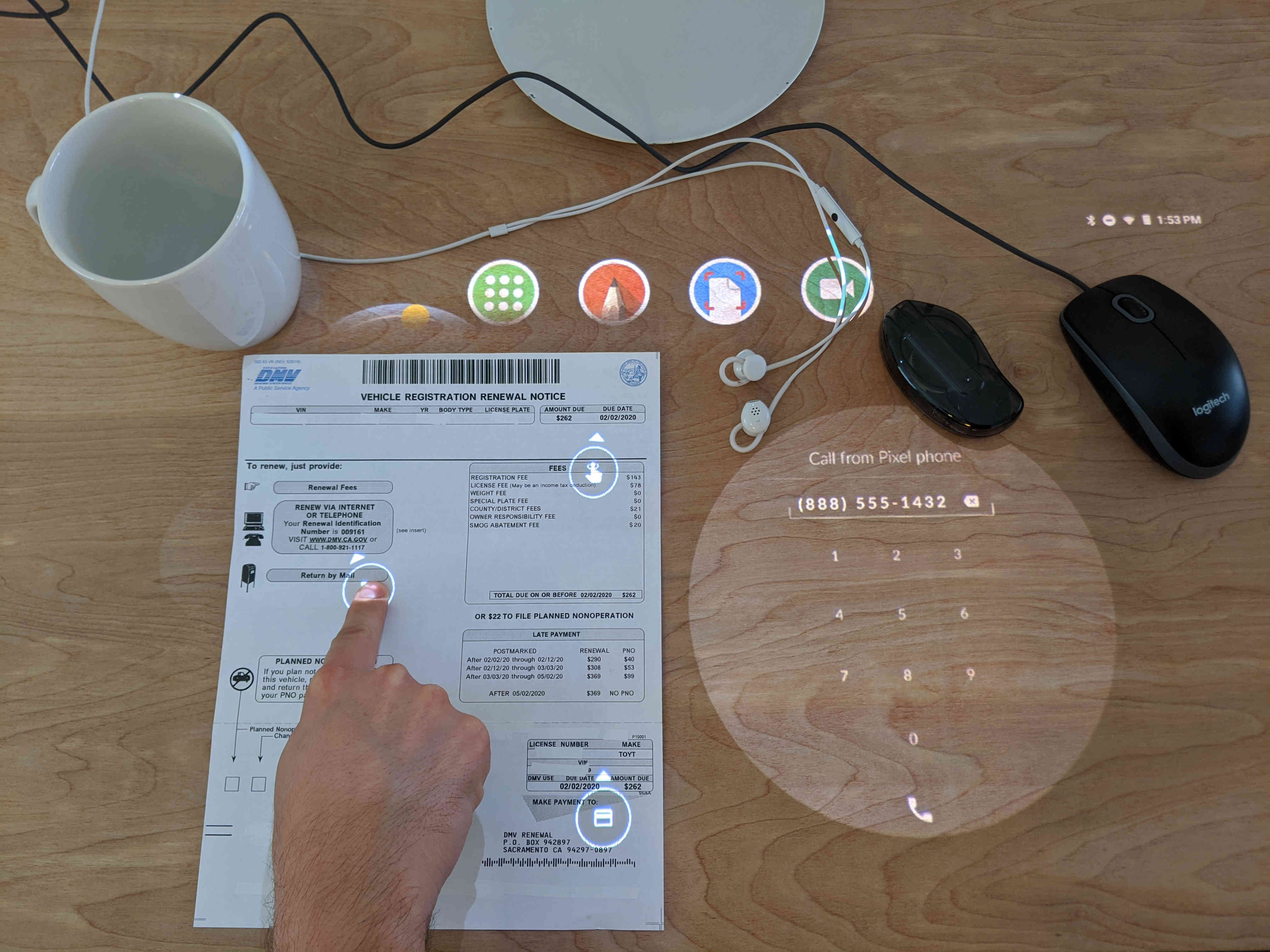}
\caption{Desktop demo application.}
\label{fig:dmv}
\end{figure}

In this demo application we explore interactions with paper documents that are on the table. 
In the past, many systems have explored these type of interactions, for example Digital Desk \cite{1993interacting} and Live Paper \cite{1999live}, project information onto physical objects.
In our application, we assume that we know the structure of the document and we attach a user interface to it, making the document interactive. The buttons allow the user to act on the information presented by the document when they are touched. 
For example, tapping a phone number could trigger the device to call the number. Tapping on the money due could trigger the device to pay the bill.
By tracking the document, using optical flow, the user can move it through the table and
have the illusion that the graphical user interface is attached to the document.
See Figure \ref{fig:dmv} for an example of a document that we can detect, track,
and project a graphical user interface on.

\section{Conclusions}
\label{sec:conclusions}

In this paper we introduced a new interactive projector-camera prototype with
a lamp form factor. The prototype has a novel real-time algorithm that can
detect touches and in-air hand gestures.
The low-cost prototype uses a middle-tier SoC that is similar to
a Google Pixel 2 smartphone and has a vision module that is similar to the RealSense D435.
The novel real-time algorithm leverages the hardware acceleration SoC
capabilities, that are present in many smartphones.
The low-cost prototype is robust to clutter  which are important
limiting factors for the adoption of interactive tabletop projectors.
We showed how the novel prototype enables rich experiences in which the
user can interact with the tabletop and objects of interests in realistic
tabletop scenarios.

\section{Acknowledgements}
We would like to thank Rahul Sukthankar and Jay Yagnik for their support and advice. We would also like to thank John Fitch, Eric Cuong Nguyen, Kai Yick, and Mark Zarich for their assistance with hardware prototyping.

\footnotesize \bibliographystyle{acm}
\bibliography{bib/hands, bib/touch}

\end{document}